# Considerations and Results in Multimedia and DVB Application Development on Philips Nexperia Platform

Radu Arsinte[1], Ciprian Ilioaei[2]

**Abstract** – This paper presents some experiments regarding applications development on high performance media processors included in Philips Nexperia Family. The PNX1302 dedicated DVB-T kit used has some limitations. Our work has succeeded to overcome these limitations and to make possible a general-purpose use of this kit. For exemplification two typical applications, important both for multimedia and DVB, are analyzed: MPEG2 video stream decoding and MP3 audio decoding. These original implementations are compared (in speed, memory requirements and costs) with Philips Nexperia Library.

## I. INTRODUCTION

Modern multimedia embedded applications are present in different forms in our life. DVB set-top boxes, DVD players, satellite receivers are few examples of this kind of well-known products.
Implementing embedded multimedia applications is possible only by using high performance processors. Using general PC based platforms for development is possible, but the goals to achieve lowest cost, lowest-consumption products are possible only by using so called "media processors".
Such processors are present in the offer of many large semiconductor companies. Some examples are given in [1], [2].
Philips, a recognized pioneer in video-audio technology is involved in development of a high-performance, low-cost media processors, Nexperia PNX1300 Series which delivers up to 200 MHz of power to a variety of multimedia applications. PNX1300 Series processors achieve over seven billion operations per second in applications requiring real-time processing of video, audio, graphics, and communications datastreams.
PNX1300 processors are ideal building blocks for devices required to process several types of multimedia datastreams simultaneously, including the latest standards such as MPEG-4, MPEG-2, H.263, MP3, and Dolby Digital®. With ample computational power available to capture, compress, and decompress many video and audio data formats in real time, PNX1300s are well suited for a broad range of applications such as Internet appliances, Web-cams, smart display pads, video and screen phones, PVR, videoconferencing, video editing, video based security, Internet radios, DVD playback, wireless LAN devices, and digital TV sets and set-top boxes. They also support applications in a JavaTM virtual machine environment.
Supported by the comprehensive TriMediaTM SDE software development environment, PNX1300s are comparable in ease of programmability to general-purpose processors. The SDE enables multimedia application development entirely in the C and C++ languages.
Our work was intended to make an exploration of the Trimedia (Nexperia) and integrate this technology into a general multimedia system development. Our previous work in DVB technology was rather theoretical [3], and this occasion, to use a high performance processor has offered the opportunity to start real-time embedded multimedia implementations.

## II. DEVELOPMENT SYSTEM-DESCRIPTION

The system used for present development was initially designed for straight DVB-T applications.
The block schematic is presented in fig.1.

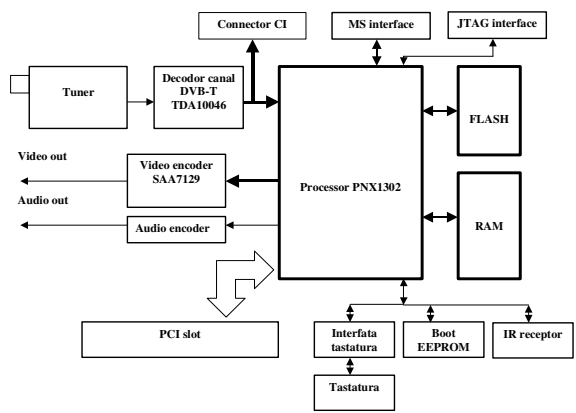

*Fig.1. Block diagram of the DVB-T Nexperia receiver*

[1] Facultatea de Electronică şi Telecomunicaţii Cluj-Napoca, Departamentul Comunicaţii, Str. G.Baritiu 26-28,
e-mail radu.arsinte@com.utcluj.ro
[2] Tedelco SRL, Calea Turzii 42, Cluj-Napoca

Few details regarding this block schematic.
1. Processor PNX1302
- offers data processing capabilities
2. Tuner
- RF processiong of incoming TV signal
- re encodes the audio/video information into RF channel
3. Channel decoder (TDA10046)
- COFDM demodulation
- outputs TS (Transport Stream) to Nexperia
4. CI Connector (Common Interface)
· links the receiver module with CI (Conditional Access) module
· Transmits a scrambled signal and receives the descrambled signal
5. Video encoder
· transforms video output stream (VO) of PNX1302 in CVBS PAL/SECAM/NTSC
· video data transfer is performed using ITU656 standard
6. Audio encoder
· Converts audio I2S in analog audio
7. PCI slot
· standard PCI interface –used to add (interface) of compatible devices
8. Flash memory
· stores the executable program
9. RAM memory
· temporary stores data and settings
10. MS (Micro Stick) interface
· used to store data  (removable peripheral)
11. Keyboard Interface
· reads local keys
12. JTAG interface - used in debugging processes

III. SOFTWARE DEVELOPMENT IN NEXPERIA ENVIRONMENT

Software support for Nexperia family has a main component IADK (Integrated Advanced Development kit). IADK contains the libraries of all the components needed in applications and the NDK (Nexperia Development Kit). We had also software support for the stand-alone systems(SAS).
The environment has the following folders:
  -audio: the libraries for the audio software components like Audio Digitizer ,Audio Renderer or MP3 Decoder.
  -video: libraries for the video components like Video Digitizer ,Video Renderer ,Mpeg decoder etc.
  -tssa: libraries for some components that make some actions like File Reader or Copy IO
  -mdm: libraries for Transport Stream Demux and Programme Stream Demux components
  -net :libraries for HTTP network communication support  and for RPC sockets
  -build: the directory where we built components file libraries and the applications for  our board.
  -sas: contains the SAS environment support.

IV. RESULTS

The development system was used to test some original applications useful in laboratory works and demonstrations. Most applications avoid the copyright problems using original implementations for algorithms and code sequences. This was a requirement of the project, making the applications independent of IADK, witch costs about 10.000$ for the smallest configuration. It is necessary to have only NDK, witch has an affordable cost.
Here are some of the applications and brief results of tests.

1.PCM Player: this application plays PCM files (*.pcm), which contain audio PCM sequences
Characteristics:
-   program was tested on and works fine
-   using non-streaming architecture
-   has a video interface, coming from YUV files
-   limitation given by the RAMDISK size
-   compiled with nohost option for our board

2.YUV Player: this application displays images that are read from the .y, .u and .v files
Characteristics:
-   program was tested on Philips ATV1 board and works good
-   using non-streaming architecture
-   no memory limitation
-   compiled with nohost option for our board

3.MP3 Player: this application plays mp3 files up to 128kb bitrate
Characteristics:
-   program was tested on Philips_ATV1
-   doesn't works in real time(for the moment)
-   limitation given by the RAMDISK size
-   compiled with nohost option for our board

4.AAC decoder/player: this application decodes and plays the encoded AAC files
Characteristics:
-   program was tested on Philips_ATV1 board
-   using non-streaming architecture
-   limitation given by RAMDISK size
-   doesn't work in real time(at this moment)
-   compiled with nohost option for our board

5.AAC encoder :this application encodes .wav and .pcm files to AAC format
Characteristics:
-   program was tested on Philips_ATV1 and works fine
-   using non-streaming architecture
-   compiled with nohost option for our board

6.AVI decoder :this application decodes uncompressed AVI  files and displays the images
Characteristics:

- program was tested on Philips_ATV1 board
- using non-streaming architecture
- limitation given by RAMDISK size
- the program works fine for the small AVI's
- compiled with nohost option for our board

7.MPEG-2 video decoder: this application implements the MPEG-2 decoding algorithm (IDCT, Huffman, etc) and displays the images
Characteristics:
- program was tested on Philips_ATV1 board and it works fine
- almost reaches real-time(93-95%)
- using non-streaming architecture
- compiled with nohost option for our board

8.Image processing: some standard operations like binarization, edge detection were implemented over a picture.

## V. CONCLUSION AND FUTURE WORK

Our activity brought us the following achievements:

1. Work with the compiler and with the other Trimedia tools.
2. Unterstanding the way makefiles work.
3. Creating the executables (*.out) for some specific applications
4. Simulating those executable files with tmsim.
5. Building the support for all given platforms: foxbox (ATV),DVE , legs
6.Generating the library files (*.a) for all the software Trimedia components that came with the two CD's.

## REFERENCES

[1] Texas Instruments - *TMS320DM641/TMS320DM640 Video/ Imaging Fixed-Point Digital Signal Processors* – Data Manual , June 2003 .
[2] ST Microelectronics - STi5518 – *Single-Chip SET-TOP BOX Decoder with MP3 and Hard Disk Drive Support* – data sheet  - 2001
[3] Radu Arsinte, Ciprian Ilioaei - *Some Aspects of Testing Process for Transport Streams in Digital Video Broadcasting* – Acta Technica Napocensis, Electronics and Telecommunications, vol.44, Number 1, 2004
[4] Radu Arsinte - *A Low Cost Transport Stream (TS) Generator Used in Digital Video Broadcasting Equipment Measurements* – Proceedings of AQTR 2004 (THETA 14) - 2004 IEEE-TTTC-International Conference on Automation, Quality and Testing,Robotics May 13-15, 2004, Cluj-Napoca, Romania